\documentclass[lettersize,journal]{IEEEtran}
\usepackage{amsmath,amsfonts}
\usepackage[noend]{algorithmic}
\usepackage{algorithm}
\usepackage{array}
\usepackage[caption=false,font=normalsize,labelfont=sf,textfont=sf]{subfig}
\usepackage{textcomp}
\usepackage{stfloats}
\usepackage{url}
\usepackage{verbatim}
\usepackage{graphicx}
\usepackage{cite}
\usepackage[table]{xcolor}
\usepackage{makecell}
\usepackage{multirow}
\DeclareMathOperator*{\argmin}{arg\,min}

\begin{document}

\title{Continual Learning by Three-Phase Consolidation}

\author{Davide Maltoni, Lorenzo Pellegrini*
\thanks{Both authors from University of Bologna, Dept. of Computer Science and Engineering - DISI.}
\thanks{E-mail: davide.maltoni@unibo.it, l.pellegrini@unibo.it}
\thanks{*Corresponding author}
}

\markboth{Preprint under review}%
{Davide Maltoni and Lorenzo Pellegrini: Continual Learning by Three-Phase Consolidation}


\maketitle

\begin{abstract}
TPC (Three-Phase Consolidation) is here introduced as a simple but effective approach to continually learn new classes (and/or instances of known classes) while controlling forgetting of previous knowledge. Each experience (a.k.a. task) is learned in three phases characterized by different rules and learning dynamics, aimed at removing the class-bias problem (due to class unbalancing) and limiting gradient-based corrections to prevent forgetting of underrepresented classes. Several experiments on complex datasets demonstrate its accuracy and efficiency advantages over competitive existing approaches. The algorithm and all the results presented in this paper are fully reproducible thanks to its publication on the Avalanche open framework for continual learning.
\end{abstract}

\begin{IEEEkeywords}
Continual Learning, Class-incremental, Memory Consolidation, Forgetting.
\end{IEEEkeywords}

\section{Introduction}
\label{sec:intro}
\IEEEPARstart{W}{hen} a neural network is trained continually, and the whole past data is no longer accessible, the model suffers from catastrophic forgetting of previous knowledge \cite{McCloskey1989}. Despite the recent introduction of a number of effective approaches, Continual Learning (CL) remains challenging on complex real-world applications with high data dimensionality, frequent updates, and stringent computation constraints. Furthermore, proposed techniques are often complex, governed by several hyper-parameters, and difficult to port to scenarios/datasets different from the native ones.

The memorization of part of past data and its successive replay (a.k.a. experience replay \cite{icarl}) was found to be one of the simplest and the most effective methods to control forgetting. At the same time, regularization techniques such as distillation \cite{LWF}, especially when combined with replay \cite{DER, BIC} provided good results in a variety of scenarios. However, distillation and replay come at a cost, which can be relevant in applications requiring frequent updates. CL in the class-incremental (and class-incremental with repetition) scenarios is characterized by experiences where only a subset of the classes is present in each experience (differently from the i.i.d. hypotheses underlying SGD) leading to the well-known class-bias problem \cite{FACIL}, where the unrepresented (or underrepresented) classes are overwhelmed by the others. Existent methods explicitly designed for bias correction are among the most effective solutions \cite{maltoni2019, BIC} but have some drawbacks (as discussed in Section 3).

In this work, we introduce TPC a simple and efficient CL approach that can be applied to complex scenarios characterized by frequent updates with small data chunks. The design choices of TPC are guided by simplicity: the algorithm does not include distillation components and/or complex mechanisms for the selection of replay samples. Therefore, TPC can be easily parametrized and ported to novel scenarios. It can work even without replay to comply with cases where past data storage is not possible (e.g. for privacy constraints). The method splits the learning of each experience (i.e., by partitioning the total number of epochs) into three phases with the aim of simultaneously controlling the class bias and previous knowledge forgetting:

\begin{itemize}
    \item The first phase is a sort of bootstrap for the novel classes (never encountered before): in fact, these classes cannot still compete with the known ones and must be raised in a “protected” environment. Otherwise, the known classes would respond stronger than the novel classes (whose weights have initially small values), and exaggerated gradient corrections would be applied to reduce their responses, leading to forgetting.
    \item In the second phase the novel classes reached a certain maturity and all the classes seen so far can be updated simultaneously. However, a relevant class unbalancing may be present in the mini-batches and some protections are necessary. Two mechanisms are put in place: (i) the first is an online bias-correction in the classification head to avoid the most-represented classes dominating the others; (ii) the second consists of limiting the gradient backpropagation only to necessary cases (e.g., a wrong class attracts a sample more than the ground truth class). Avoiding unnecessary gradient updates better preserves previous knowledge.
    \item A final consolidation is performed in the third phase to reach optimal equilibrium, where sample-balancing is enforced for all the classes seen so far so that the resulting model becomes totally class-neutral.
\end{itemize}

Selectively blocking gradient down-propagation in phase I and II helps in reducing forgetting and its role is very significant when replay cannot be used. The proposed online bias correction was found to be more effective than post-experience correction (adopted by CWR \cite{maltoni2019}, BiC \cite{BIC}) since the model remains bias-free during the learning of the current experience, leading to better optimization of the class boundaries.

TPC was compared with competitive existing approaches on several complex scenarios achieving better accuracy and demonstrating a good accuracy/efficiency tradeoff. Most of the TPC hyperparameters are non-critical and were fixed across the experiments; this is an important property given the recognized difficulty of running CL algorithms on datasets/scenarios different from the native ones.

Even if we avoid any speculative comparison between TPC and memory consolidations in biological systems \cite{Klinzing2019, 10.7554/eLife.51005, 10.1162/neco_a_01433}, it is known that memory consolidation during sleep takes place in a number of stages (Light Sleep, Slow-Wave Sleep, Rapid Eye Movement) characterized by different dynamics where hippocampus and cortex works together to fixate important details while limiting forgetting. This a complex multi-objective task requiring the learning of totally new concepts, relating them to existing ones (to point out similarities and allow discrimination), avoiding interfering with unrelated ones, and, finally, optimizing the overall storage subject to capacity constraints: doing all the above through a single homogeneous phase was probably out of reach for evolution and (in our opinion) for current CL algorithms.

In Section \ref{sec:notation} we introduce the notation and the problem definition. In Section \ref{sec:related} we review related literature and point out the novelty of the proposed approach. Section \ref{sec:tpc} provides a detailed description of TPC, including the underlying class bias correction and gradient masking. In Section \ref{sec:experiments_all} we discuss the datasets and algorithms used for our experimental validations and provide several results, including an ablation study to better point out the contribution of the method building blocks. Some concluding remarks are finally drawn in Section \ref{sec:conclusions}.

\section{Notation and Problem definition} 
\label{sec:notation}
A continual learning problem consists of a number $n_e$ of experiences $E_i$ (a.k.a. tasks\footnote{The term \textit{task} was introduced in CL literature for scenarios where disjoint tasks need to be learned sequentially by a model. In single incremental task scenarios (as class incremental) the term experience looks more appropriate and less ambiguous. Furthermore, experience is the term adopted by the Avalanche framework used in this paper.}), each containing a subset of data that is only accessible during the corresponding training stage:

\begin{equation}
\textit{CL} = \{ E_1,E_2,\ldots E_{n_e} \}
\end{equation}

\begin{equation}
E_i\!=\!(E_i^x,E_i^y),E_i^x\!=\!\{x_1^i,x_2^i,\ldots x_{n_i}^i\},E_i^y\!=\!\{y_1^i,y_2^i,\ldots y_{n_i}^i\}
\end{equation}

where $E_i^x$ and $E_i^y$ are the datapoints and the associated labels contained in the $i$-th experience and $n_i$ is the number of samples in the $i$-th experience.

Depending on the distribution of labels in the experiences, different scenarios can be defined (see Section 2 of \cite{Negative} for a formal definition). For example, in the class-incremental scenario (addressed by most recent studies), each $E_k$ contains only classes never seen in previous experiences:

\begin{equation}
\label{eq:eq1}
E_k^y \cap E_i^y = \emptyset, \forall i<k
\end{equation}

Here we relax constraint \ref{eq:eq1} and allow each experience to contain both samples of novel classes and new samples of classes encountered in previous experiences. This scenario known as class-incremental with repetitions (or NIC) is closer to real applications (see \cite{CIR}), where the observation of novel instances of previous classes (a sort of natural replay) can improve robustness and better cope with data drift. 
Therefore each $E_i = (E_i^x,E_i^y ) = (E_i^{x_{novel}} \cup E_i^{x_{rep}},E_i^{y_{novel}} \cup E_i^{y_{rep}})$ is composed of samples of novel classes and new samples of some of the known classes (repetitions). In general, the classes in $E_i^{y_{rep}}$ are a subset of all the classes seen so far (denoted as known classes). It is worth noting that in the extreme cases where $E_i^{y_{rep}} = \emptyset$ or  $E_i^{y_{novel}} = \emptyset$ we fall back in the class-incremental and domain-incremental scenario, respectively.  
 Given a classification model $f$ parametrized by $\Theta$, fitting it incrementally through the sequence of experiences, by simply continuing SGD to minimize cross-entropy loss is prone to forgetting:

\begin{equation}
\label{eq:eq2}
\Theta^* = \argmin_{\Theta} \mathcal{L}_{CE} (f_\Theta (E_i^x), E_i^y) \quad for\ i = 1 \ldots n_e
\end{equation}

To protect the old knowledge the loss function can be extended by including regularization components (e.g., constraining the value of critical parameters as pioneered by EWC \cite{EWC} and/or adding distillation terms, first introduced by LWF \cite{LWF}). Furthermore, storing into a memory $\mathcal{R}$ (of fixed capacity $n_{replay}$) a subset of past data and replaying them jointly with the samples of current experience was demonstrated to be very effective. In the case of replay equation \ref{eq:eq2} becomes:

\begin{equation}
\label{eq:eq3}
\Theta^* = \argmin_{\Theta} \mathcal{L}_{CE} (f_\Theta (E_i^x \cup \mathcal{R}_i^x), E_i^y \cup \mathcal{R}_i^y) \; for\ i = 1 \ldots n_e
\end{equation}

where $\mathcal{R}_i^x$ and $\mathcal{R}_i^y$ are the datapoints and labels contained in the replay memory at the beginning of experience $E_i$. The replay memory is updated at the end of each experience by adding a subset of samples from the current experience. Note that repetition and replay samples are different: the former are novel instances or known classes while the latter are samples already encountered in the past\footnote{Actually replay samples can change (across different presentations) if augmentation is applied after sampling them from the memory. However, unlike repetition samples, they are derived by from already seen samples.}.

The model $f_\Theta$ considered in this work is a classifier based on a convolutional neural network. Starting from the input layer we can divide the model into three blocks of layers such that $f_\Theta$ is their functional composition:

\begin{equation}
\label{eq:eq4}
f_\Theta = c_{\Theta_c} \circ {csf}_{\Theta_{csf}} \circ {llf}_{\Theta_{llf}}
\end{equation}

where:

\begin{itemize}
    \item ${llf}$ consists of a block of layers extracting Low-Level Features parametrized by $\Theta_{llf}$; such features are quite generic and portable throughout homogeneous applications (e.g., natural image classification).
	\item ${csf}$ includes layers working on top of ${llf}$ and rearranging features to extract Class-Specific Features. Such features (and corresponding parameters $\Theta_{csf}$) are shared among classes. 
	\item $c$ is the final Classification head (a linear dense layers spanning $n_{classes}$) whose parameters $\Theta_c$ can be partitioned into disjoint groups of class-exclusive parameters $\Theta_{c_j}, j=1 \ldots n_{classes}$.
\end{itemize}

While the identification of $c$ in the model architecture is straightforward (i.e., the single last layer of the model), the boundary between $llf$ and $csf$ is arbitrary and can be application application-specific. The proposed architecture splitting is often beneficial in the practice when using models pre-retrained on large datasets, but is it not a pre-requisite of TPC: in fact, without loss of generality, $llf$ can be set to $\emptyset$ and all the layers (except the final head) associated to $csf$.

\section{Related works} 
\label{sec:related}

Several continual learning approaches have been proposed in the last decade. Comprehensive surveys have been published to categorize methods and summarize their pros and cons \cite{FACIL, Parisi2019, MAI202228}. Hereafter we avoid re-proposing a generic introduction to CL and focus on methods and techniques more related to TPC.

\subsection{Bias correction} 
\label{sec:bias_correction}
Bias-correction methods are among the most effective CL techniques available for class-incremental scenarios (see survey by Masana et al. \cite{FACIL}) and a number of methods have been proposed for class-bias correction in the last layer \cite{lesort2022continual, BIC, WA, Rebalancing, caccia2022new}.

CWR, which is an essential component of AR1 \cite{maltoni2019} and ARR \cite{ARR} algorithms, is one of the first approaches introduced for bias correction. Since the bias correction approach designed for TPC was inspired by CWR here after we provide more details on it.

CWR corrects the class bias in the classification head by learning the novel classes in isolation and performing a posterior weight normalization and consolidation. Learning in isolation limits the forgetting of the known classes that are not represented in the current experience and leaves the novel classes free to learn without competing with known classes whose weights are already mature. To learn in isolation the novel classes, CWR maintains a copy $\Theta_c^{'}$ of the weights of the classification head $\Theta_c$ of the previous experience: at the beginning of each experience the classification head weights $\Theta_c$ are reset and only weights of classes of the current experience are loaded from $\Theta_c^{'}$. At the end of the experience, the weight consolidation phase takes place, where the weights $\Theta_c$ learned in the current experience are consolidated with past weights $\Theta_c^{'}$. In particular, for each parameter group $\Theta_{c_j}$ associated with a class $j$ belonging to the current experience $E_i$, the mean of the group weights $\mu_j=avg(\Theta_{c_j} )$ is calculated, and subtracted to all the weights in the group, in order to force zero mean: $\Theta_{c_j}=\Theta_{c_j}-\mu_j$. This was demonstrated to be effective to prevent class-bias due to the different magnitudes of the weights. Consolidation is then performed by weighted average:

\begin{equation}
\label{eq:eq5}
\Theta_{c_j} = \frac{{w_{past_j} \cdot \Theta_{c_j}^{'}} + \Theta_{c_j}}{w_{past_j} + 1}
\end{equation}

where $w_{{past}_j}$ is the weight of the past and can be computed according to the ratio between the number of samples $n_{{past}_j}$ seen in the past for class $j$ and its number of samples $n_{{cur}_j}$ in $E_i$. However, the definition of $w_{{past}_j}$ can be critical for a long sequence of experiences where the contribution of the current experience tends to vanish over time. In \cite{Rehearsal_free}, in order to reduce this vanishing effect a square root is used instead of a simple proportion, leading to the CWR* variant:

\begin{equation}
\label{eq:eq6}
w_{{past}_j}=\sqrt{n_{{past}_j}/n_{{cur}_j}}
\end{equation}

One of the most effective bias correction approaches is BiC \cite{BIC} which is often taking the top positions in the evaluations reported in recent papers. This method, which is considered in our comparative evaluations (Section \ref{sec:experiments_all}), is based on distillation and bias correction. For distillation, a second model is maintained (as a memory of the past), and while the new model is learned a distillation loss component enforces a certain stability in the responses of the two models. The bias correction mechanism requires isolating a validation set from the training data of the current experience which is used just after the experience training to learn two scaling parameters for each class to balance the magnitude of the responses. Analogously to CWR, BiC needs to define the importance of the past, in order to properly weigh the distillation components of the loss, and therefore is potentially prone to the above mentioned learning vanishing effect. Furthermore, BiC cannot work without replay and needs to sacrifice a small part of training data for internal validation. BiC is also computationally more complex than TPC because of the need to perform inference on two models.

\subsection{Gradient masking} 
\label{sec:gradient_masking_related}
Gradient masking is a technique to block (or reduce) the weight corrections that a gradient descent algorithm would apply. It can be applied by selectively resetting (or reducing) some elements of the gradient vector during the backward pass. In principle, the loss function itself could be modified to achieve similar goals, but in our experience with gradient masking is simpler to focus on specific changes without introducing side effects. For example, hinge loss could be adopted to avoid penalizing small errors (one of the goals in TPC phase II); however, in our experiments, we found it less effective than Cross Entropy + Gradient masking.

Gradient masking was used to limit forgetting in the output layer in \cite{lesort2022continual} whose authors propose two masking techniques: (i) \textit{single masking} only updates weights for the output vector of the true class; (ii) \textit{group masking} masks all classes that are not in the mini-batch. The gradient masking carried out in TPC differs from both the above techniques: more details are provided in Section \ref{sec:gradient_masking}. Furthermore in \cite{lesort2022continual} gradient masking is applied only to the classification head since all the remaining layers were frozen after pre-training.

Gradient masking was also applied in the CL approach \cite{Negative} where the backpropagation of real and generated datapoints is dealt with differently to leverage generated data as negative examples when learning new classes without allowing them to change the knowledge of old classes.

\section{Three-Phase Consolidation} 
\label{sec:tpc}
The TPC algorithm is introduced in section \ref{sec:tpc_algorithm} after a description of the two basic mechanisms on which it relies: online bias correction and gradient masking.

\subsection{Online bias correction} 
\label{sec:online_bc}
TPC includes a novel online approach for class-bias correction in the classification head $c$. It was inspired by CWR normalization of the weight groups, but has relevant differences and advantages:

\begin{itemize}
    \item CWR performs the correction at the end of each experience by normalizing (to zero mean) each group of weights $\Theta_{c_j},  j=1\ldots n_{classes}$. On the contrary, TPC correction is on-line and therefore weights are always updated and immediately contribute to the optimization of the class boundaries. Furthermore, there is no need to maintain a double classification head. 
	\item for the classes of the current experience that were already encountered before (i.e., they belong to $E_i^{y_{rep}}\cup\mathcal{R}_i^y$), CWR at the end of the experience consolidates the weights according to a weighted average keeping into account the importance of the past (see equation \ref{eq:eq5}). This can be critical (and difficult to parametrize) for long continual learning sequences because the importance of the past constantly increases making the updates progressively less effective. On the contrary, in TPC weights $\Theta_c$ are always kept updated and there is no need to setup and parametrize any a-posteriori consolidation.
    \item In CWR the classes included in the current experience are learned in isolation (based on a properly initialized temporary head). Learning in isolation has some advantages (see discussion in Section \ref{sec:bias_correction}) but does not allow to effectively discriminate between the classes in the experience and old classes (non-represented in the experience).
\end{itemize}

TPC bias correction can be performed in two ways: by explicit normalization or KL loss extension.

\begin{enumerate}
    \item \textbf{Explicit normalization}: is performed after each optimizer step (i.e. once the gradient correction has been applied for the current mini-batch) by forcing each group of weights $\Theta_{c_j}$ to zero mean and fixed standard deviation ($s$):
    \begin{align}
    \label{eq:eq7}
 \Theta_{c_j} &= s \cdot \frac{\Theta_{c_j}-\mu_j}{\sigma_j},\\ \notag
 where\ \mu_j = avg&\left(\Theta_{c_j}\right)\ and\ \sigma_j=std\left(\Theta_{c_j}\right)
    \end{align}
    The classes considered by the normalization are all the classes seen so far (including those of the current experience). We experimentally found that setting $s<1$ (we used $0.05$ in our experiments) provides better results probably due to the extra regularization enabled by smaller weight values. We also noted that imposing a fixed standard deviation is here essential, while in CWR a simple mean shift is enough.
    \item \textbf{BC loss extension}: the normalization of the weight groups can be performed by adding a term $\mathcal{L}_{BC}$ to the cross-entropy loss $\mathcal{L}_{CE}$. $\mathcal{L}_{BC}$, which is based on KL divergence, pushes the weights in each group $\Theta_{c_j}$ to follow a normal distribution $\mathcal{N}(0,s)$:
    \begin{equation}
    \label{eq:eq8}
    \mathcal{L}_{BC}\!=\!\!\frac{1}{2\!\cdot\! n_{classes}}\!\!\!\!\sum_{j=1}^{n_{classes}}\!\!\left[\!\left(\!\frac{\mu_j}{s}\!\right)^{\!2}\!\!\!+\!\!\left(\!\frac{\sigma_j}{s}\!\right)^{\!2}\!\!\!-\!log\!\left(\!\!\left(\!\frac{\sigma_j}{s}\!\right)^{\!2}\!\!\!+\!\epsilon\!\right)\!\!-\!1\!\right]
    \end{equation}
    where \textit{eps} is a small constant inserted to avoid numerical problems. The idea is similar to the method used by Variational Autoencoders to constrain the distributions of the latent variables $z$ \cite{KingmaW13}, and its derivation for $\mathcal{N}(0,1)$ can be found in Appendix B of \cite{KingmaW13}. The extension to $\mathcal{N}(0,s)$ leading to equation \ref{eq:eq8} can be derived by the KL-Divergence between two normal distributions (see details in Appendix \ref{sec:appendix_kl}). $\mathcal{L}_{BC}$ can be added to the cross entropy loss by including a weighting factor ($w_{BC}$) to balance the importance of the two components. It is worth noting that $w_{BC}$ is a fixed parameter (non-critical to tune) and does not depend on the importance of the past.
    \begin{equation}
    \label{eq:eq9}
    \mathcal{L}=\mathcal{L}_{CE}+w_{BC}\cdot\mathcal{L}_{BC}
    \end{equation}
\end{enumerate}

Explicit normalization has the advantage of an immediate and exact correction of the bias after each iteration, but leads to a zig-zag optimization due to the alternation of disjoint gradient corrections and post-normalization steps. BC loss extension determines a smoother optimization due to the simultaneous minimization of cross-entropy and class bias, and performed slightly better in our experiments, so it was adopted as the default strategy in TPC. To remove the small residual bias that could be present at the end of the experience training a single explicit normalization step is added at the end of phase III (see Algorithm \ref{alg:tpc}).

\subsection{Gradient masking} 
\label{sec:gradient_masking}

Given a classification head $c$ with $n_{classes}$ output neurons and a minibatch of size $n_{mb}$ of samples, then the gradient $G$ sent back across this layer during backpropagation is a 2D tensor (i.e. a matrix) with the same shape of the output predictions, that is ($n_{mb}$, $n_{classes}$). We can selectively block the error backpropagation by resetting one or more values of this tensor. In particular, we can set to 0 the entire column corresponding to class $k$, if we do not want to backpropagate any corrections for that class\footnote{It is worth noting that blocking corrections through the output neuron corresponding to a given class does not make the model completely stable with respect to that class, since the shared weights $\Theta_{csf}$ (if not frozen) can be changed through the corrections sent back via output neurons of other classes.}. Alternatively, we can select some examples in the minibatch (i.e., selecting some rows) and set to 0 some specific columns only for those examples: this provides maximum flexibility since we can decide which samples are allowed to backpropagate corrections for which classes. Gradient masking is applied differently depending on the TPC phase:

\begin{itemize}
    \item In phase I, which is a sort of bootstrapping for classes $E_i^{y_{novel}}$ in the classification head, we want to avoid the corrections to other classes. In fact, until the weights of novel classes reach a certain strength, other classes could respond strongly and their corrections lead to forgetting. Therefore, we mask the gradient for all classes $k\notin E_i^{y_{novel}}$ (see Table \ref{tab:table1} for an example). It is worth noting that such masking is different than training the model only on examples belonging to $E_i^{y_{novel}}$ since with the proposed masking all the examples in the minibatch (including those belonging to $E_i^{y_{rep}}$ and $\mathcal{R}_i^y$) contribute to modify the weights of $E_i^{y_{novel}}$ in $\Theta_c$, and samples belonging to $E_i^{y_{novel}}$ have no impact on the weights in $\Theta_c$ of other classes. This also differs from the group masking proposed in \cite{lesort2022continual} that masks all classes not in the mini-batch, while we mask also some classes in the mini-batch, namely $E_i^{y_{rep}}\cup \mathcal{R}_i^y$.
    \begin{table}[ht]
\centering
\begin{tabular}{|l|c|c|c|c|c|}
\hline
 & \multicolumn{5}{c|}{\textbf{Predicted class}} \\ \hline
\textbf{True Class} & \textbf{1} & \textbf{2} & \textbf{3} & \textbf{4} & \textbf{5} \\ \hline
$4 \in E_i^{y_{novel}}$ & \cellcolor{lightgray}0.30 & \cellcolor{lightgray}0.20 & \cellcolor{lightgray}0.30 & 0.15 & \cellcolor{lightgray}0.05 \\ \hline
$4 \in E_i^{y_{novel}}$ & \cellcolor{lightgray}0.20 & \cellcolor{lightgray}0.30 & \cellcolor{lightgray}0.20 & 0.25 & \cellcolor{lightgray}0.05 \\ \hline
$2 \in E_i^{y_{rep}}$ & \cellcolor{lightgray}0.10 & \cellcolor{lightgray}0.70 & \cellcolor{lightgray}0.10 & 0.05 & \cellcolor{lightgray}0.05 \\ \hline
$2 \in E_i^{y_{rep}}$ & \cellcolor{lightgray}0.05 & \cellcolor{lightgray}0.70 & \cellcolor{lightgray}0.10 & 0.10 & \cellcolor{lightgray}0.05 \\ \hline
$1 \in\mathcal{R}_i^y$ & \cellcolor{lightgray}0.65 & \cellcolor{lightgray}0.15 & \cellcolor{lightgray}0.10 & 0.05 & \cellcolor{lightgray}0.05 \\ \hline
$2 \in\mathcal{R}_i^y$ & \cellcolor{lightgray}0.05 & \cellcolor{lightgray}0.80 & \cellcolor{lightgray}0.05 & 0.05 & \cellcolor{lightgray}0.05 \\ \hline
$3 \in\mathcal{R}_i^y$ & \cellcolor{lightgray}0.10 & \cellcolor{lightgray}0.10 & \cellcolor{lightgray}0.65 & 0.10 & \cellcolor{lightgray}0.05 \\ \hline
\end{tabular}
\captionsetup{justification=justified}
\caption{In this example, we show the activation tensor corresponding to a mini-batch containing 7 datapoints: the first 2 belonging to the novel class 4, while the remaining 2 + 3 belong to repetition and replay classes, respectively. Class 5 is a future class. Activations are reported in the cells. During phase I, the gradient of all but novel classes (only class 4 in this example) is masked (gray cells).}
\label{tab:table1}
\end{table}
    \item In phase II, the weights of all classes have reached a certain maturity. However, some classes have no (or fewer) positive examples in the mini-batches (for example the classes belonging to $\mathcal{R}_i^y$), and the risk is their excessive penalization by negative corrections imposed by examples belonging to well-represented classes. In order to limit negative corrections, we mask the gradient for the classes $k\notin E_i^{y_i}$ for the samples where the response for $k$ is small with respect to the response for the ground truth class. Ideally, when cross-entropy is applied with one-hot targets any non-zero prediction for a wrong class must be pushed toward zero: what we propose here is to tolerate small activations for wrong classes if they are not dangerous (e.g., there is no risk for misclassification). This reduces corrections for classes that cannot “defend” themselves because of underrepresentation. Formally, we mask classes $k\notin E_i^y$ for samples $\left(x,y\right)\in mb$ such that $f_\Theta(x)\left[k\right]<t\cdot f_\Theta(x)\left[y\right]$, where $f_\Theta(x)\left[k\right]$ denotes the activation of class $k$, $f_\Theta(x)\left[y\right]$ is the activation of the ground truth class $y$, and $t$ is a threshold (see Table \ref{tab:table2} for an example). Again, this differs from the single masking proposed in \cite{lesort2022continual} where the gradient is backpropagated only for the ground true class, while we retropropagate the error also for other classes when there is a risk of misclassification.
    \begin{table}[ht]
\centering
\begin{tabular}{|l|c|c|c|c|c|}
\hline
& \multicolumn{5}{c|}{\textbf{Predicted class}} \\ \hline
\textbf{True Class} & \textbf{1} & \textbf{2} & \textbf{3} & \textbf{4} & \textbf{5} \\ \hline
$4 \in E_i^{y_{novel}}$ & \cellcolor{lightgray}0.15 & 0.15 & 0.25 & 0.40 & \cellcolor{lightgray}0.05 \\ \hline
$4 \in E_i^{y_{novel}}$ & 0.30 & 0.05 & \cellcolor{lightgray}0.10 & 0.50 & \cellcolor{lightgray}0.05 \\ \hline
$2 \in E_i^{y_{rep}}$ & \cellcolor{lightgray}0.10 & 0.60 & \cellcolor{lightgray}0.20 & 0.05 & \cellcolor{lightgray}0.05 \\ \hline
$2 \in E_i^{y_{rep}}$ & \cellcolor{lightgray}0.05 & 0.70 & \cellcolor{lightgray}0.10 & 0.10 & \cellcolor{lightgray}0.05 \\ \hline
$1 \in\mathcal{R}_i^y$ & 0.70 & 0.10 & \cellcolor{lightgray}0.10 & 0.05 & \cellcolor{lightgray}0.05 \\ \hline
$2 \in\mathcal{R}_i^y$ & 0.30 & 0.55 & \cellcolor{lightgray}0.05 & 0.05 & \cellcolor{lightgray}0.05 \\ \hline
$3 \in\mathcal{R}_i^y$ & \cellcolor{lightgray}0.10 & 0.10 & 0.65 & 0.10 & \cellcolor{lightgray}0.05 \\ \hline
\end{tabular}
\captionsetup{justification=justified}
\caption{The same mini-batch of Table \ref{tab:table1} is processed during Phase II. Here we do not mask classes 2 and 4 because they belong to $E_i^y$. The gradient of the remaining classes (1, 3, and 5) is selectively masked (gray) when the activations are small (less than $t=50\%$ with respect to the true class activation).}
\label{tab:table2}
\end{table}
    \item In phase III classes are balanced and masking is not necessary.
\end{itemize}

\subsection{The TPC algorithm} 
\label{sec:tpc_algorithm}

The pseudocode of TPC is provided in Algorithm \ref{alg:tpc} for the default case when the model is pretrained and a limited replay memory can be used. In our opinion, this setup is the most effective to solve most of the real-world applications.

\begin{algorithm*}
\caption{TPC Pseudocode}
\label{alg:tpc}
\begin{algorithmic}
\STATE freeze $\Theta_{llf}$   \COMMENT{see main text for the case where the model is not pretrained}
\STATE $\mathcal{R}=\emptyset$ \COMMENT{initialize replay memory}
\STATE $known=\emptyset$ \COMMENT{initialize known classes}
\STATE Loss $\mathcal{L}=\mathcal{L}_{CE}+w_{BC}\mathcal{L}_{BC}$ \COMMENT{the same loss across all phases (Eq. \ref{eq:eq9})}
\FOR{each training experience $E_i$}
\STATE $\Theta_{c_j}=0,$ for each class $j, j \notin known$
\STATE $y_{novel}=E_i^y-known$
\STATE \COMMENT{Phase I}
\IF{$i>1$}
\STATE freeze $\Theta_{csf}$ \COMMENT{there is no forgetting in the first experience}
\ENDIF
\FOR{epoch in ${epochs}_1$}
\FOR[the number of iterations is $n_i/n_{mbe}$]{iterations}
\STATE $mb \gets$ load $n_{mbe}$ datapoints from $(E_i^x,E_i^y)$ and $n_{mbr}$ datapoints from $(\mathcal{R}_i^x,\mathcal{R}_i^y)$
\STATE SGD step with gradient masking of classes $k\notin E_i^{y_{novel}}$ for all samples $(x,y)\in  mb$
\ENDFOR
\ENDFOR
\STATE \COMMENT{Phase II}
\STATE unfreeze $\Theta_{csf}$
\FOR{epoch in ${epochs}_2$}
\FOR[the number of iterations is $n_i/n_{mbe}$]{iterations}
\STATE $mb \gets$ load $n_{mbe}$ datapoints from $(E_i^x,E_i^y)$ and $n_{mbr}$ datapoints from $(\mathcal{R}_i^x,\mathcal{R}_i^y)$
\STATE SGD step with gradient masking of classes $k\notin E_i^y$ for samples $(x,y)\in mb$ such that $f_\Theta(x)[k]<t \cdot f_\Theta(x)[y]$
\ENDFOR
\ENDFOR
\STATE \COMMENT{Phase III}
\STATE Update $\mathcal{R}$ from $E_i$ with class-balanced reservoir sampling
\FOR{epoch in ${epochs}_3$}
\FOR[the number of iterations is $n_{replay}/n_{mbe}$]{iterations}
\STATE $mb \gets$ load $n_{mbe}$ datapoints from $(\mathcal{R}_i^x,\mathcal{R}_i^y)$
\STATE SGD step
\ENDFOR
\ENDFOR
\STATE Explicit normalization step (Eq. \ref{eq:eq7}) \COMMENT{to remove any residual class bias}
\STATE $known=known \cup E_i^y$
\ENDFOR
\end{algorithmic}
\end{algorithm*}

Since the model was pre-trained low-level features are assumed to be portable and $\Theta_{llf}$ are kept frozen. The three phases are clearly identifiable: phases I and III typically run for 1 or very few epochs while Phase II is the main one and takes all the rest of epochs. In Phase I, while novel classes are bootstrapped, the shared parameters $\Theta_{csf}$ are kept frozen, to avoid undesired corrections for other classes. On the contrary, in phases II and III $\Theta_{csf}$ are free. The extended loss (with the BC term) is constant throughout all the phases, while gradient masking is different: in phase I it blocks gradient propagation for all but the novel classes while in phase II it allows backpropagating updates for all classes while blocking only corrections for small activations of wrong classes. The replay memory is populated by class-balanced reservoir sampling (see \cite{reservoir, Isele_Cosgun_2018}) which is a simple but effective approach to balance classes and experience representation in $\mathcal{R}$. In phases I and II, the minibatch $mb$ consists of $n_{mbe}$ datapoints from the current experience and $n_{mbr}$ datapoints from the replay memory, while in phase III only the (updated) replay memory is used. While in phases I and II datapoints from current experience are visited once per epoch, the replay memory can be only partially accessed or can be re-visited more than once depending on the number of iterations ($n_i/n_{mbe}$), the replay size $n_{replay}$ and $n_{mbr}$. A discussion on how to set an optimal proportion between $n_{mbr}$ and $n_{mbe}$ is proposed in Section \ref{sec:experiments}. Augmentation is applied by default to datapoints from current experience and replay memory. Latent replay (that is storing in $\mathcal{R}$ activations just after $llf$ instead of raw datapoints see \cite{latent}) is also feasible and computationally very efficient, even if data augmentation for replay data is no longer possible; in this paper, we do not use latent replay.

Minor variants of the same algorithm can be set to let TPC work without replay memory or without model pre-training:

\begin{itemize}
    \item If data for replay cannot be stored, letting $\Theta_{csf}$ free is dangerous in the absence of other protection mechanisms (e.g., weight protection by Synaptic Intelligence \cite{Zenke2017} approach such as in \cite{maltoni2019}) which are not easy to make working in real scenarios with many experiences. However, if the model was pre-trained on a similar domain, a reasonable approach is keeping $\Theta_{llf}$ frozen, tuning $\Theta_{csf}$ in the first experience and learning only the classification head weights $\Theta_c$ in all the successive experiences. This is demonstrated with specific experiments on Core50 in Section \ref{sec:ablation}.
    \item If the model needs to be trained from scratch, the first experience is hopefully large enough to bootstrap the feature extraction layers. In this case, we learn $\Theta_{llf}$ only during $E_1$ (frozen for the rest of the experiences). Specific experiments on Cifar100 are reported in Section \ref{sec:results}.
\end{itemize}

Finally, the case of no pre-training and no replay memory was sometimes addressed in the literature (see for examples AR1 approach in \cite{Rehearsal_free}) but we believe it is too difficult for the complex dataset/scenarios addressed in this paper, so we do not recommend using TPC in such cases.

\section{Experiments} 
\label{sec:experiments_all}

\subsection{Datasets} 
\label{sec:datasets}

The benchmarks considered in our experiments focus on challenging scenarios, where the datasets Core50 \cite{core50} and ImageNet1000 \cite{imagenet} are split into a large number of experiences (at least 40). Cifar100 \cite{cifar} is also used to test learning-from-scratch capabilities. Hereafter, according to \cite{FACIL} we adopt the naming convention: Dataset \textit{A/B-C}, where $A$ is the total number of experiences ($n_e$ in Section \ref{sec:notation}), $B$ is the number of classes included in the first experience ($E_1$) and $C$ is the number of classes in each of the successive experiences $E_i,i>1$. In the following we discuss the setting adopted in the experiments.

Class incremental with a pre-trained model:
\begin{itemize}
	\item \textbf{Core50 41/10-1}: Core50 \cite{core50} is an object recognition dataset specifically introduced for continual learning. The classical setup used for Core50 in the class incremental scenario is Core50 9/10-5. Here we raise the complexity by increasing the number of experiences to 41. Note that all the experiences, but the first, include examples of a single new class; this could be required in practical applications where the model needs to be updated each time a new class is encountered to be able to immediately predict it. The CNN model used is a variant of the MobileNet v1 \cite{Howard2012} (more details in Appendix \ref{sec:appendix_model}, ImageNet pre-trained, input size 128x128) with the first 20 blocks (20 convolutional layers) in $llf$ and the last 7 blocks (7 convolutional layers) in $csf$. The capacity of the replay memory was set to $n_{replay}=1500$ according to common practices on Core50 \cite{latent}.
	\item \textbf{ImageNet1000 100/10-10}: The ImageNet dataset when used full size (1000 classes, 1.3 M samples, and full image resolution) is a quite challenging setup for continual learning. In fact, it is often used in the mini- or tiny- version (see \cite{miniimagenet, chaudhry2019tiny} and \cite{tinyimagenet, DER, MasanaT021}). The settings used in the CL literature for ImageNet1000 often include a moderate number of experiences (max 25). Here too we raise the complexity and propose a benchmark with 100 experiences of 10 classes each. The CNN model used is a ResNet-18 \cite{resnet} (pretrained on Places365 \cite{places365}, image size 224x224) with the first 3 blocks (13 convolutional layers) in $llf$ and the last block (4 convolutional layers) layers in $csf$. The capacity of the replay memory was set to $n_{replay}=20000$ according to common practices on ImageNet1000 \cite{FACIL}.
\end{itemize}

Class incremental without a pre-trained model:

\begin{itemize}
    \item \textbf{Cifar100 11/50-5}: we adopt one of the most common Cifar100 setting among those proposed in the literature. This dataset is used in our experiment to test “continual learning from scratch”, that is without using a pre-trained model. In our opinion, this is not a good setup to work on natural images, but it could be necessary in specific domains where no transferable pre-trained models are available. The CNN model used is a ResNet-32 (as the one used in \cite{icarl, FACIL}) with the first 2 blocks (21 convolutional layers) in $llf$ and the last block (10 convolutional layers) in $csf$. The capacity of the replay memory was set to $n_{replay}=2000$ according to common practices on Cifar100 \cite{FACIL}.
\end{itemize}

Class incremental with repetitions:

\begin{itemize}
    \item \textbf{Core50 NICv2 391/10-1}: this is one of the few class-incremental with repetition scenarios available for testing continual learning algorithms \cite{CIR}. Each of the 390 experiences (after the first one) contains samples of at most one class $c$, that can either belong to $E_i^{y_{novel}}$ or $E_i^{y_{rep}}$. Furthermore, the number of samples in each experience $E_i, i>1$  is quite small (about 300), making incremental learning challenging. The CNN model used is the same as the one used in Core50 41/10-1. The capacity of the replay memory was set to $n_{replay}=1500$ according to common practices on Core50 NICv2 \cite{latent}.
\end{itemize}

\subsection{Algorithms} 
\label{sec:algorithms}

TPC is compared with 3 other algorithms: AR1, BiC, and DER++. These approaches were selected taking into account: (i) accuracy (state-of-the-art or close to state-of-the-art); (ii) efficiency (we cannot run too complex methods on computing expensive scenarios), (iii) maturity, and (iv) availability. Good performance reported in more (independent) studies is an indicator of maturity. All the algorithms selected have public implementation (by the authors or third parties) made available in Avalanche framework \cite{Avalanche, JMLR:v24:23-0130}, allowing us to run and compare all the methods in exactly the same setting (see Section \ref{sec:experiments}). TPC implementation will be also made available in Avalanche for full reproducibility of the experiments and further studies.

\begin{itemize}
	\item \textbf{AR1}: AR1 (see \cite{ARR} for a comprehensive introduction), was proved to be an effective CL technique, overcoming classical continual learning approaches on several benchmarks. Furthermore, it relies on CWR which is a strong bias correction approach. The version here considered does not include Synaptic Intelligence \cite{Zenke2017} protection of weights in $csf$ because that component is quite critical for CL over a large number of sequences, whereas using a replay memory proved to be more effective (denoted as AR1free in \cite{latent}).
	\item \textbf{BiC}: BiC \cite{BIC}, already discussed in Section \ref{sec:bias_correction}, is recognized as one of the best-performing CL algorithms (see conclusions in the survey by Masana et al. \cite{FACIL}). BiC is a two phases approach: the first phase consists of a training step performed using an LwF-like \cite{LWF} distillation approach\footnote{For distillation an old model is maintained and, at each experience, samples are forwarded through both the models.}; during the second phase, a pair of scaling parameters ($\alpha$ and $\beta$) are learned for the current classes and used to de-bias the output logits ($y=\alpha o+\beta$). The existing Avalanche implementation of BiC was based on the code made available by authors of \cite{FACIL}, where the scaling parameters learned during the bias-correction phase were permanently incorporated into the model. We noted that such an implementation strategy based on permanent scaling layers is different (and underperforming) with respect to the original code released by BiC authors so we provided an Avalanche implementation of BiC which is aligned with the native one (more details can be found in Appendix \ref{sec:appendix_bic}). 
	\item \textbf{DER++}: DER and its variant DER++ \cite{DER} store replay examples along with their “soft labels” (logits). In plain DER, a knowledge distillation loss component is added with the aim of minimizing the difference between the current and past responses for the replay examples. This loss component is weighted with a strength factor $\alpha$. In DER++, a more classic classification loss component is also added that uses the ground truth label of replay examples. This loss component is weighted with a strength factor $\beta$ (with $\beta=0$, DER++ collapses to DER). The DER++ idea is simple, but the approach is considered a strong baseline because it performed reasonably well on several setups.
\end{itemize}

We are not considering classical approaches such as LwF \cite{LWF}, EwC \cite{EWC}, Icarl \cite{icarl}, Gem \cite{lopez2017gradient} and the plethora of their variants, because they usually struggle in complex class-incremental settings where the selected approaches were proved to perform better by several researchers. 

\subsection{Experimental setup} 
\label{sec:experiments}

The idiom “the devil is in the detail” particularly fits CL experiments, where apparently insignificant changes have a relevant impact on the results. For example, training details such as total iterations, learning rate schedule, data augmentation, amount of replay, and the proportion of replay data in the mini-batches, have an impact on the accuracy-efficiency tradeoff. Publications on CL rarely report the efficiency of the tested methods and this makes comparison critical. Therefore, we decided to run all the selected approaches exactly on the same conditions and hardware platform, by relying on Avalanche \cite{Avalanche} facilities. In particular, all the methods use the same: (i) CNN architecture and pre-training; (ii) optimizer (SGD with momentum) (iii) data augmentation; (iv) amount of replay; (v) mini-batch size and composition in terms of $n_{mbr}$ and $n_{mbe}$; (vi) number of epochs. In particular the ratio $n_{mbe}/n_{mbr}$ in a mini-batch was set as the ratio between the number of samples $n_s$ in each experience $E_i, i>1$  and the total size of the replay buffer $n_{replay}$:
\begin{equation}
n_{mbe}:n_{mbr}=n_s:n_{replay}
\end{equation}

This choice (introduced by \cite{latent}) is quite optimal in many scenarios.
Finally, the specific hyperparameters of each method (plus the learning rate) have been coarsely tuned on the first run of each experiment and kept constant for the rest of the runs. With coarse-tuning we mean that we started with the default values suggested in the original papers (or in public implementations) and tuned them with a coarse grid search (more details are provided in Appendix \ref{sec:appendix_setup}). It is worth noting that for TPC we kept the hyperparameters $s$, $w_{BC}$ and $t$ constant across all the experiments\footnote{The proportion among the number of epochs associated with the three phases was also kept constant.}. While this could lead to a small accuracy drop in our comparative evaluation, we believe it can simplify TPC porting to new setups.

\subsection{Results} 
\label{sec:results}

\begin{figure*}[!t]
\centering
\subfloat{
    \includegraphics[width=2.7in]{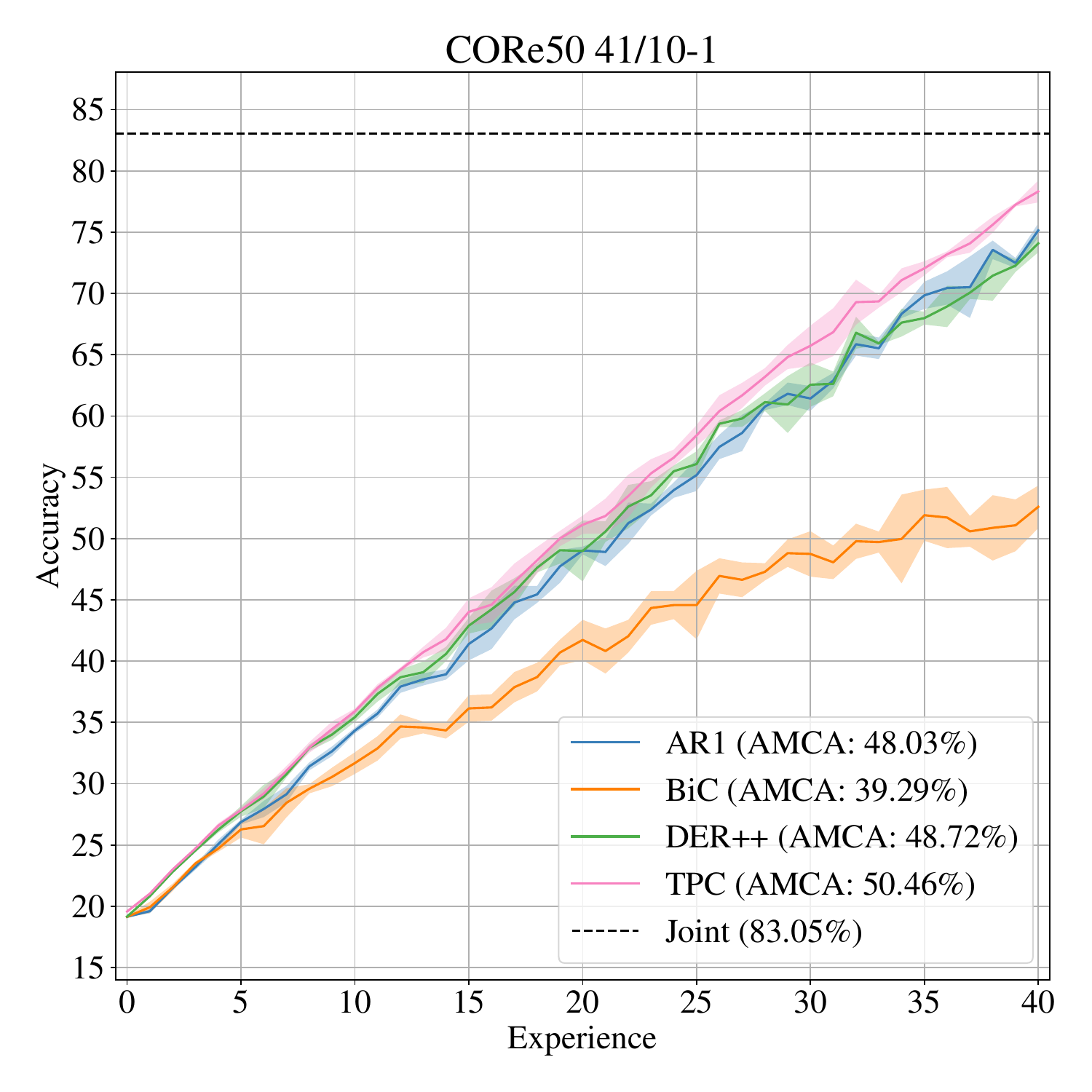}%
    \label{fig:results_core50_41_10_1}
}
\hfil
\subfloat{
    \includegraphics[width=2.7in]{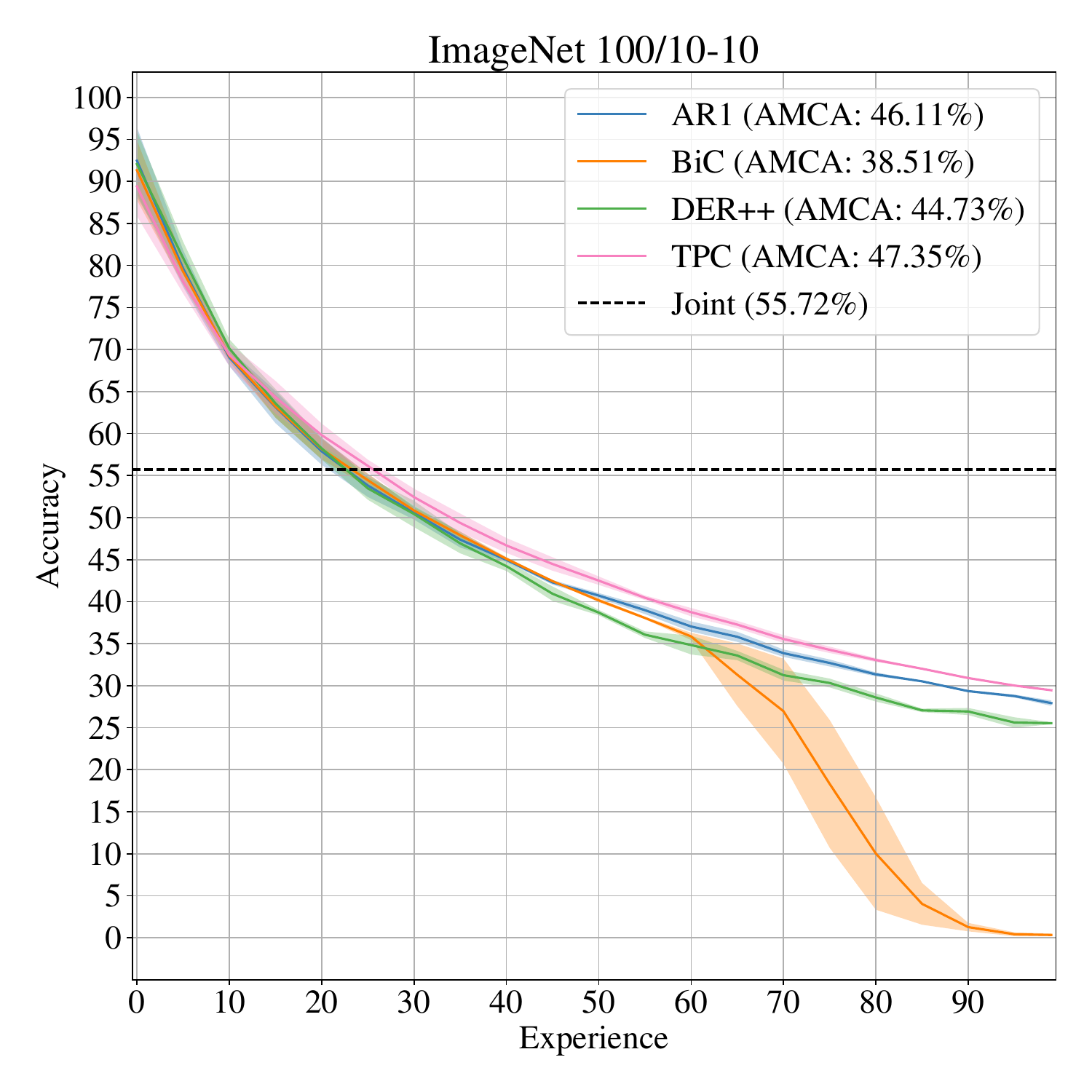}%
    \label{fig:results_imagenet}
}\\
\subfloat{
    \includegraphics[width=2.7in]{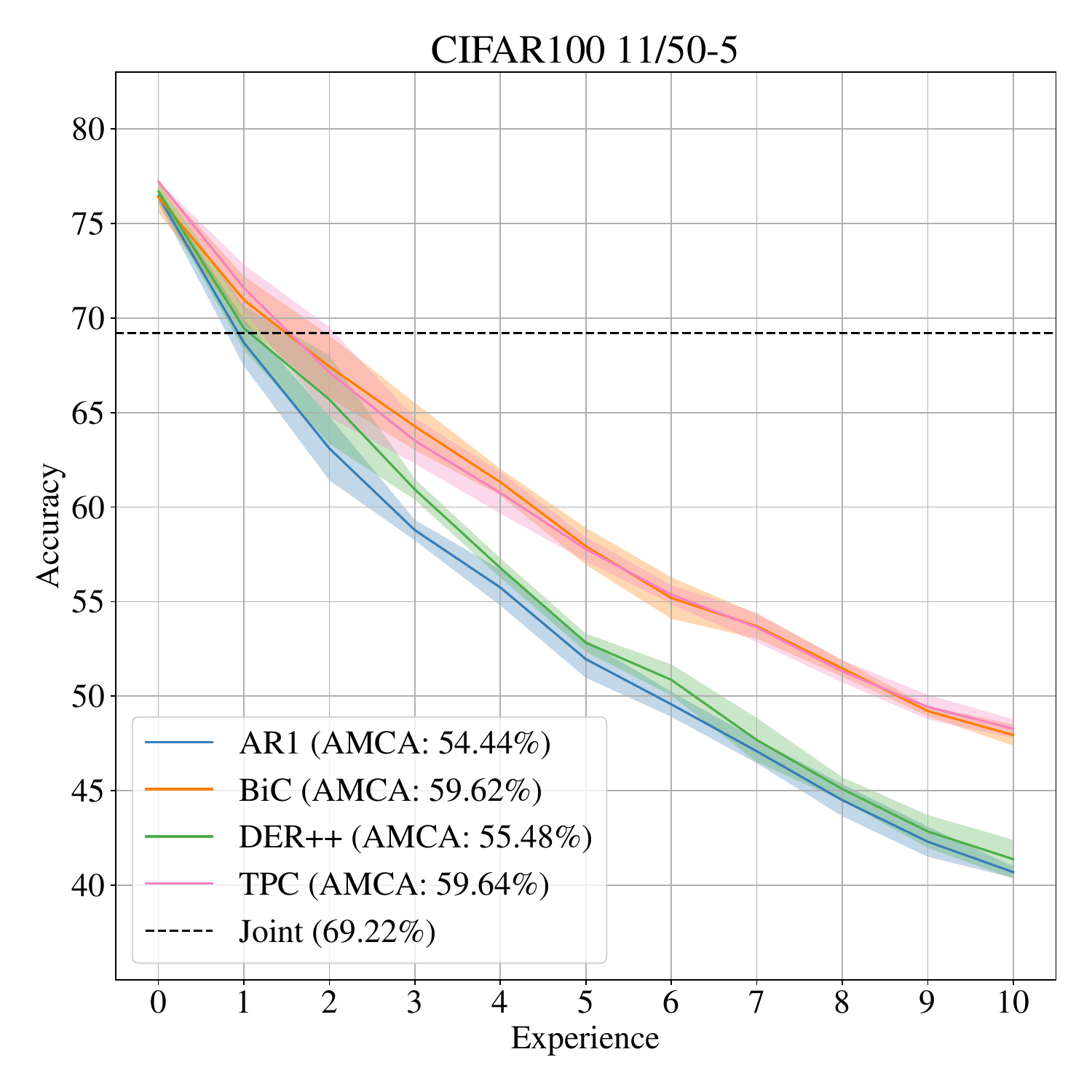}%
    \label{fig:results_cifar}
}
\hfil
\subfloat{
    \includegraphics[width=2.7in]{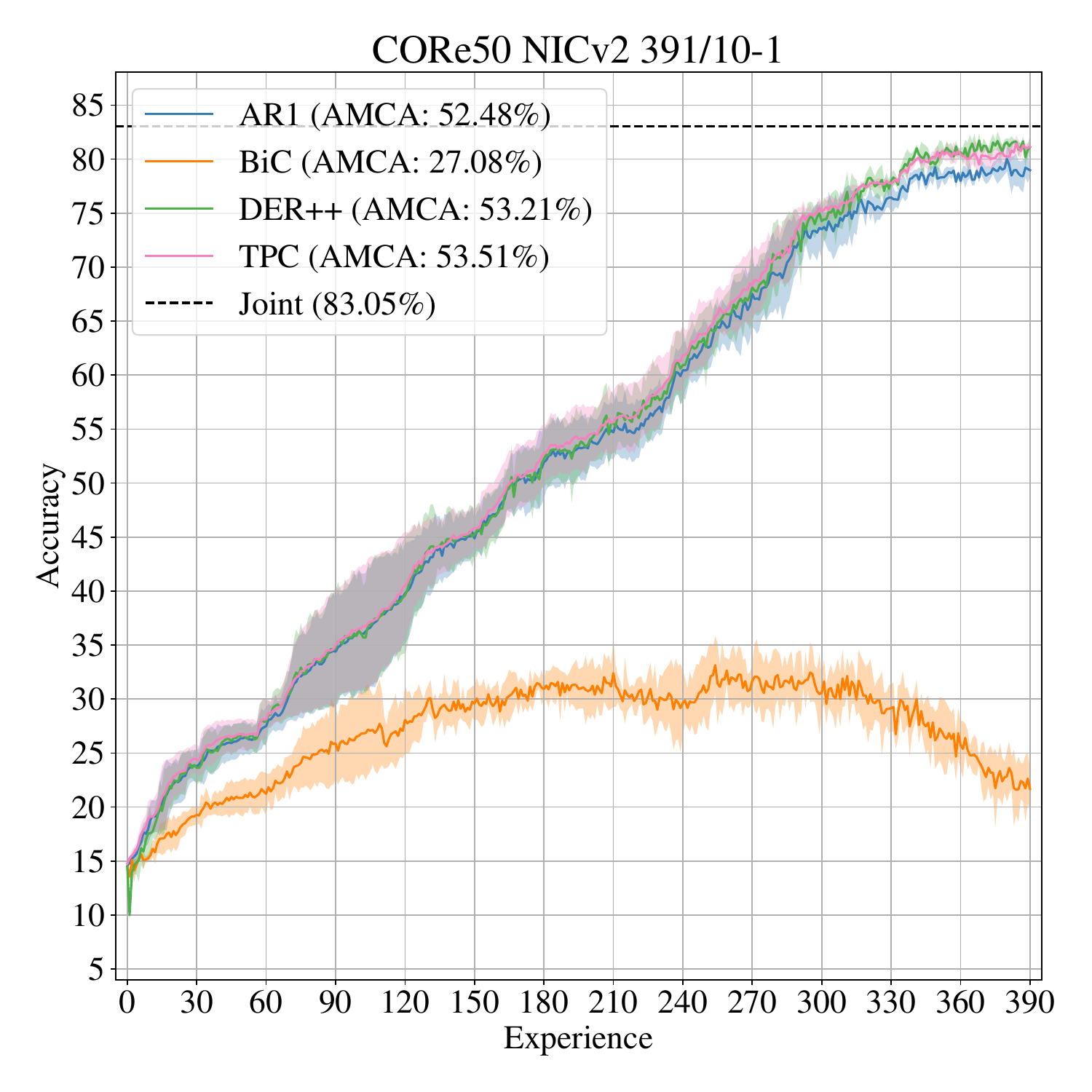}%
    \label{fig:results_core50_nic}
}

\caption{Accuracy on the four benchmarks. Each curve is the average over 3 runs (with different ordering of the classes); the run used for hyperparams tuning is excluded. As a common practice in the existing studies, on Core50 benchmarks accuracy is measured on a fixed test set where all classes (also those not yet learned) are present (see \cite{core50}), while on ImageNet1000 100/10-10 and Cifar100 11/50-5 is measured on an incremental test set including only the classes seen so far. The dashed line (which represents a sort of upper bound) denotes the accuracy of the same model jointly trained on all the data.}
\label{fig:results}
\end{figure*}

Figure \ref{fig:results} shows the results on the four benchmarks. Tables \ref{tab:table_amca}, \ref{tab:table_final_accuracy}, and \ref{tab:table_times} report: the overall AMCA (Average Mean Class Accuracy over the experiences), the final accuracy (after last experience), and the training time, respectively.

\begin{table*}[ht]
\centering
\begin{tabular}{l|l|l|l|l|l}
\textbf{AMCA $\uparrow$} & CORe50 41/10-1 & ImageNet 100/10-10 & CIFAR100 11/50-5 & CORe50 NICv2 391/10-1 & Overall \\ \hline
AR1 & 48.03\% (0.95) & 46.11\% (0.97) & 54.44\% (0.91) & 52.48\% (0.98) & 0.95 \\
BiC & 39.29\% (0.78) & 38.51\% (0.81) & 59.62\% (1.00) & 27.08\% (0.51) & 0.78 \\
DER++ & 48.72\% (0.97) & 44.73\% (0.94) & 55.48\% (0.93) & 53.21\% (0.99) & 0.96 \\
TPC (ours) & \textbf{50.46\% (1.00)} & \textbf{47.35\% (1.00)} & \textbf{59.64\% (1.00)} & \textbf{53.51\% (1.00)} & \textbf{1.00} \\
\end{tabular}
\caption{For each benchmark and approach, we report the AMCA. Within brackets and in the last column (overall) the accuracy is reported as multiple of TPC accuracy.}
\label{tab:table_amca}
\end{table*}
\begin{table*}[ht]
\centering
\begin{tabular}{l|l|l|l|l|l}
\textbf{Final accuracy $\uparrow$} & CORe50 41/10-1 & ImageNet 100/10-10 & CIFAR100 11/50-5 & CORe50 NICv2 391/10-1 & Overall \\ \hline
AR1 & 75.14\% (0.96) & 27.92\% (0.95) & 40.70\% (0.84) & 78.97\% (0.97) & 0.93 \\
BiC & 52.59\% (0.67) & 0.34\% (0.01) & 47.94\% (0.99) & 21.67\% (0.27) & 0.49 \\
DER++ & 74.08\% (0.95) & 25.53\% (0.87) & 41.38\% (0.86) & 81.11\% (1.00) & 0.92 \\
TPC (ours) & \textbf{78.32\% (1.00)} & \textbf{29.46\% (1.00)} & \textbf{48.28\% (1.00)} & \textbf{81.14\% (1.00)} & \textbf{1.00} \\
\end{tabular}
\caption{For each benchmark and approach, we report the final accuracy. Within brackets and in the last column (overall) the accuracy is reported as multiple of TPC accuracy.}
\label{tab:table_final_accuracy}
\end{table*}
\begin{table*}[ht]
\centering
\begin{tabular}{l|l|l|l|l|l}
\textbf{Time $\downarrow$} & CORe50 41/10-1 & ImageNet 100/10-10 & CIFAR100 11/50-5 & CORe50 NICv2 391/10-1 & Overall \\ \hline
AR1 & \textbf{257 s (0.97)} & 83135 s (1.01) & \textbf{1797 s (0.95)} & \textbf{1224 s (0.77)} & \textbf{0.93} \\
BiC & 572 s (2.16) & 137746 s (1.67) & 3494 s (1.85) & 2827 s (1.77) & 1.86 \\
DER++ & 558 s (2.11) & 83104 s (1.01) & 3170 s (1.68) & 3031 s (1.90) & 1.68 \\
TPC (ours) & 265 s (1.00) & \textbf{82606 s (1.00)} & 1892 s (1.00) & 1599 s (1.00) & 1.00 \\
\end{tabular}
\caption{For each benchmark and approach, we report the training time measured on the same HW (See Appendix \ref{sec:appendix_timing}). Within brackets and in the last column (overall) the training time is reported as multiple of TPC time.}
\label{tab:table_times}
\end{table*}

In the first two benchmarks (Core50 41/10-1 and ImageNet1000 100/10-10) TPC emerges as the best performing approach. In particular, the accuracy of TPC on Core50 41/10-1 is also very stable (i.e., without oscillations) across experiences, and the final accuracy is quite close to the joint training upper bound\footnote{Joint training refers to an experiment where the model is trained on the entire dataset without incurring incremental learning burdens.}. AR1 and DER++ are not far from TPC while BiC struggles to work with such a large number of experiences including few classes. In particular, BiC performs quite well in the first half of ImageNet1000 100/10-10 experiments but it shows a consistent drop in the second part\footnote{This is consistent with other results reported on ImageNet for BIC where, if the number of experiences is limited ($\le 25$), the method is highly competitive.}.

On Cifar100 11/50-5 a smaller non-pretrained model is used with a large first experience (half of the dataset) for the boosting of low-level features. Here TPC and BiC achieve the best results, while the accuracy of AR1 and DER++ is substantially lower.

Finally, on the class incremental with repetition benchmark, TPC and DER++ achieved the best performance, AR1 follows quite closely while BiC performs poorly. It is worth noting that BiC was not designed to work on class-incremental with repetitions scenario, and to make it compatible with such scenario we introduced a simple modification where repetition classes in the current experience are considered (i) old classes: in this case, no bias correction is computed for them, or (ii) novel classes: hence subject to bias correction. Variant (ii) performed slightly better and was used to produce the reported results.

If we look at the training times, we note that AR1 and TPC are more efficient than DER++ and BiC, mainly because of the distillation, which requires further computation.

\begin{figure}[!t]
\centering
\includegraphics[width=3in]{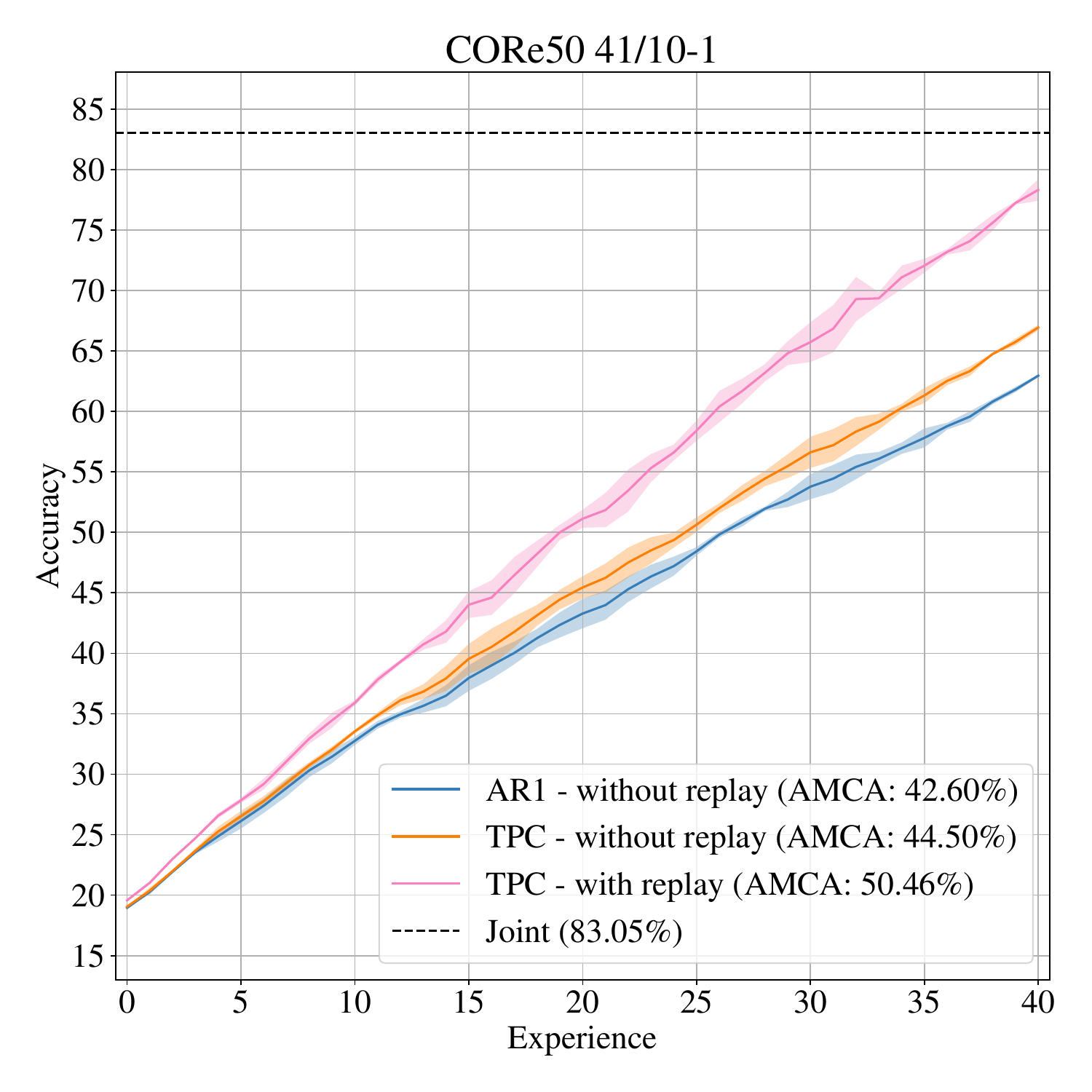}
\caption{Accuracy on Core50 41/10-1. TPC baseline (violet) is here compared with TPC and AR1 running without replay memory. AMCA is reported in the legend.}
\label{fig:no_replay}
\end{figure}

A final experiment is reported in Figure \ref{fig:no_replay} where TPC works without replay memory. As discussed in Section \ref{sec:tpc_algorithm}, to avoid forgetting in the representation part of the model, only the classification head is trained across the experiences. While a certain accuracy drop is here unavoidable, this result proves that, differently from other methods (such as BiC and DER++), TPC and AR1 can work even without replay, and TPC remains competitive with respect to AR1.

Summarizing, on the five benchmarks reported TPC always reaches top accuracy while remaining computationally lighter than BiC and DER and just slightly more complex than AR1.

\subsection{Ablation study} 
\label{sec:ablation}

The extra experiments reported in this section are aimed at understanding the contribution of the main TPC building blocks. Results are here reported for Core50 41/10-1 when the algorithm is run with and without replay memory. Table \ref{tab:table_ablation} shows the AMCA of the full version of TPC compared with partial versions.

While the full version is always the best performing, the contribution of bias correction and gradient masking are more relevant in the second row where replay samples are not used to mitigate forgetting. In general, we noted that reducing the replay buffer size increases the importance of bias correction and gradient masking.

\begin{table*}[ht]
\centering
\begin{tabular}{l|l|l|l|l}
\textbf{AMCA $\uparrow$} & Full algorithm & \makecell{No bias correction \\ (Section \ref{sec:online_bc})} & \makecell{No gradient masking \\ (Section \ref{sec:gradient_masking})} & No phase III \\ \hline
CORe50 41/10-1 & \textbf{50.46\%} & 48.08\% & 50.35\% & 49.58\% \\
CORe50 41/10-1 (No replay) & \textbf{44.50\%} & 14.63\% & 32.22\% & - \\
\end{tabular}
\caption{The AMCA of the full version of TPC compared with that of partial versions.}
\label{tab:table_ablation}
\end{table*}

\section{Conclusions} 
\label{sec:conclusions}

In this paper, we introduced a novel approach for continual learning that can deal with complex scenarios including long sequences of small experiences. The algorithm natively supports class incremental with repetition and, if necessary, can work without replay memory. The simplicity and facility of porting to new scenarios were central in the design of TPC and therefore we avoided including extra steps (such as distillation \cite{LWF, BIC, DER}, optimized selection of replay examples \cite{icarl, MIR}) that could bring a small additional accuracy increase\footnote{In some experiments, we noted that adding a naïve distillation can bring to TPC an extra $1\%$ accuracy.} in spite of extra computation and more difficult adaptability to different scenarios. For the same reason, the main TPC hypermeters were kept fixed across the experiments (as detailed in Section \ref{sec:appendix_setup}).

The comparison with other well-known CL approaches shows that TPC is competitive in terms of both accuracy and efficiency. The availability of TPC in the Avalanche framework will allow easy reproducibility of our work and easy comparability with other methods. In the future we plan to evolve TPC to work in scenarios where part of the data are unlabeled (partially unsupervised continual learning \cite{LUMP, continual_unsupervised_representation, unsupervised_progressive}).

\section*{Acknowledgments}
This research did not receive any specific grant from funding agencies in the public, commercial, or not-for-profit sectors.

\appendices
\section{KL Divergence} 
\label{sec:appendix_kl}

The KL (Kullback–Leibler) divergence between two normal distributions $P:\mathcal{N}(\mu_1, \sigma_1^2)$ and $Q:\mathcal{N}(\mu_2, \sigma_2^2)$ can be proved \cite{klproof} to be:

\begin{equation}
KL\left[P\parallel Q\right]=\frac{1}{2}\left[\frac{\left(\mu_2-\mu_1\right)^2}{\sigma_2^2}+\frac{\sigma_1^2}{\sigma_2^2}-log\frac{\sigma_1^2}{\sigma_2^2}-1\right]
\end{equation}

We assume that the distribution $P_j$ of weights in each group $\Theta_{c_j}$ (with mean $\mu_j$ and standard deviation $\sigma_j$) is normal, that is $P_j:\mathcal{N}(\mu_j, \sigma_j^2)$. If $Q:\mathcal{N}(0,s)$ is the desired target distribution, then:
\begin{equation}
KL\left[P_j\parallel Q\right]=\frac{1}{2}\left[\left(\frac{\mu_j}{s}\right)^2+\left(\frac{\sigma_j}{s}\right)^2-log\left(\frac{\sigma_j}{s}\right)^2-1\right]
\end{equation}

Loss $\mathcal{L}_{BC}$ (Eq. \ref{eq:eq9}) can be obtained by averaging KL contributions of all classes:
\begin{equation}
\mathcal{L}_{BC}\!=\!\!\frac{1}{2\!\cdot\! n_{classes}}\!\!\!\!\sum_{j=1}^{n_{classes}}\!\!\left[\!\left(\!\frac{\mu_j}{s}\!\right)^{\!2}\!\!\!+\!\!\left(\!\frac{\sigma_j}{s}\!\right)^{\!2}\!\!\!-\!log\!\left(\!\!\left(\!\frac{\sigma_j}{s}\!\right)^{\!2}\!\!\!+\!\epsilon\!\right)\!\!-\!1\!\right]
\end{equation}
where a small term $\epsilon$ is included to avoid numerical problems.

\section{BiC Implementation} 
\label{sec:appendix_bic}

While evaluating BiC, we found that its performance in various benchmarks (mostly the CIFAR100- and ImageNet-based ones) was inferior to those reported in the original paper. This prompted an analysis of the algorithm implementation in both Avalanche \cite{Avalanche} and FACIL \cite{FACIL}. When comparing these implementations with the original code from the authors of BiC\footnote{\url{https://github.com/wuyuebupt/LargeScaleIncrementalLearning}} (based on TensorFlow 1.x) we found an interesting discrepancy: in the Avalanche/FACIL implementations, the bias-correction scaling parameters learned in the second phase of each experience are permanently glued atop of the model to correct the output logits for the classes encountered in that experience. On the other hand, in the original implementation, the scaling parameters for the classes learned in experience $E_t$ are discarded when the second phase of the experience $E_{t+1}$ begins. In fact, such scaling parameters are implicitly and softly integrated into the model via distillation, when obtaining the logits from the frozen “old model” used for distillation in the first phase of the experience $E_{t+1}$. This behavior is clear in the author’s code, but difficult to catch from the paper. Finally, the bias correction parameters are not learned for the classes in the first experience.
A new version of BiC is released with this paper. This version was also merged into Avalanche and it is available from version 0.5.0 and later.

\section{Experimental setup details}  
\label{sec:appendix_setup}

To fairly evaluate the performance of all the algorithms compared in Section \ref{sec:experiments_all}, we considered 4 runs for each benchmark, where each run differs by the order of the classes presented in the benchmark. We used the first run to tune the hyperparameters of each algorithm (according to a grid search maximizing the AMCA metric), while the remaining 3 runs were used to compute the average accuracy. In the experiments reported in Section \ref{sec:experiments_all}, only the results of these 3 runs are reported.

For all the algorithms, we kept the model architecture, number of epochs, batch size, learning rate scheduling strategy, and augmentation procedure constant.
For the AR1 and TPC strategies, the only hyperparameter tuned is the learning rate. Other TPC hypermeters ($w_{BC}$, $t$, $s$, and the ratio of epochs between different phases) were kept fixed across the experiments.
For DER++ and BiC, additional hyperparameters were tuned, as we noted that they significantly affect the results:

\begin{itemize}
	\item For DER++, we tuned $\alpha$ and $\beta$ by using the values suggested in “Appendix G: Hyperparameter Search” of the original DER++ paper: $\alpha \in [0.1, 0.2, 0.5]$ and $\beta \in [0.5, 1.0]$.
	\item For BiC, we tuned the number of the training epochs for phase 2 while a constant number of epochs is used for the main training phase.
\end{itemize}

Table \ref{tab:table_shared_hparams} provides details of the hyperparameters that are common across all algorithms, while Table \ref{tab:table_specific_hparams} describes the algorithm-specific ones.

\begin{table*}[ht]
\centering
\begin{tabular}{|l|p{2.1cm}|p{2.4cm}|p{4.4cm}|p{3.0cm}|}
\hline
 & CORe50 41/10-1 & ImageNet 100/10-10 & CIFAR100 11/50-5 & CORe50 NICv2 391/10-1 \\ \hline
Epochs for $E_1$ & 4 & 35 & 200 & 4 \\ \hline
Epochs for $E_i,\ i=2\ldots n_e$ & 4 & 35 & 150 & 4 \\ \hline
Learning rate for $E_1$ & \multirow{4}{=}[\normalbaselineskip]{[0.05, 0.05, 0.005, 0.002, 0.0005]} & \multirow{4}{=}[\normalbaselineskip]{[0.01, 0.005, 0.001]} & 0.1 (with patience scheduler, as in \cite{FACIL}) & \multirow{4}{=}[\normalbaselineskip]{[0.05, 0.02, 0.005, 0.002, 0.0005]} \\ \cline{1-1} \cline{4-4}
Learning rate  for  $E_i,\ i=2\ldots n_e$ &  &  & [0.1, 0.05, 0.01, 0.005, 0.001, 0.0005] &  \\ \hline
$n_{mbe}$ for $E_1$ & \multicolumn{4}{c|}{128} \\ \hline
${n_{mbe}:n}_{mbr}$ for $E_i,\ i=2\ldots n_e$ & 158:98 & 50:78 & 142:114 & 42:214 \\ \hline
\end{tabular}
\caption{Hyperparameters shared by all algorithms. While elements such as the number of epochs or the batch size are fixed, the learning rate used when training on incremental experiences is searched among the given values.}
\label{tab:table_shared_hparams}
\end{table*}

\begin{table*}[ht]
\centering
\begin{tabular}{|l|p{3.5cm}|p{3cm}|p{3cm}|p{3cm}|}
\hline
 & CORe50 41/10-1 & ImageNet 100/10-10 & CIFAR100 11/50-5 & CORe50 NICv2 391/10-1 \\ \hline
AR1 & \multicolumn{4}{p{12.8cm}|}{$w_{\text{past}}$ strategy: square root} \\ \hline
BiC & Number of stage 2 epochs: [4, 8, 12] & Number of stage 2 epochs: [20, 35, 50] & Number of stage 2 epochs: [50, 100, 150] & Number of stage 2 epochs: [4, 8, 12] \\ \hline
DER++ & \multicolumn{4}{p{12.8cm}|}{$\alpha$: [0.1, 0.2, 0.5] \newline $\beta$: [0.5, 1.0]} \\ \hline
TPC & \multicolumn{4}{p{12.8cm}|}{$w_{\text{BC}}$: 5.0 \newline $t$: 0.5 \newline $s$: 0.05\newline  phase I epochs: 10\% of total training epochs \newline phase III epochs: 10\% of total training epochs} \\ \hline
\end{tabular}
\caption{Algorithm-specific hyperparameters. For AR1 and TPC, only the learning rate has been tuned (as reported in Table \ref{tab:table_shared_hparams}) while other parameters are kept fixed. For CORe50-based benchmarks, the resulting number of epochs for phases I and III of TPC are rounded up to 1 due to the low number of training epochs (4).}
\label{tab:table_specific_hparams}
\end{table*}

For instance, for the CORe50 41/10-1 benchmark we run a number of training runs equal to the number of hyperparameter combinations: AR1: $|LR| = 5$, TPC: $|LR| = 5$, BiC: $|LR| \cdot |\textit{phase 2 experiences}| = 5 \cdot 3 = 15$, DER: $|LR| \cdot |\alpha| \cdot |\beta| = 5 \cdot 3 \cdot 2 = 30$, totalling AR1 + TPC + BiC + DER = 55 grid search runs. A similar computation can be done for all the remaining benchmarks, leading to a total of 209 hyperparameter search runs. In addition, we run each algorithm on the 3 test runs for a grand total of 257 experiments.

For ablation tests with replay, we re-use the hyperparameters combination found in the main grid search and we report the results on the 3 test runs for each element in the ablation experiment table. For the no-replay ablation experiments, we re-executed the entire pipeline (grid search on full TPC, plus 3-run averaging experiments for each element in Table \ref{tab:table_ablation}).

To speed up the computation, on ImageNet 100/10-10 we only evaluate the performance on the test set at a 5 experiences interval (that is, at experience $i=[0, 5, \ldots, 95]$ plus the final one $i=99$). For other experiments, we evaluate the test accuracy after each experience, even for CORe50 NICv2 391/10-1.

All the experiments were performed by using Avalanche framework \cite{Avalanche}.

\subsection{Timing}  
\label{sec:appendix_timing}

To extract timing, we executed each algorithm without interfering elements such as metrics, logging, and evaluation phases. Times were taken on an idle system by dropping system caches or other interfering elements.

The hardware used is a workstation equipped with an AMD EPYC 7282 CPU and Quadro RTX 5000 GPU. The same number of data loading workers was used across all algorithms, taking care of using the same number of workers for the BiC phase 2 and DER++ feature extraction steps.

\subsection{Model details}  
\label{sec:appendix_model}

For the ImageNet-based benchmark, a standard ResNet-18 pretrained on Places365 has been used with an input image size of 224x224.

For the CIFAR100-based benchmark, a randomly initialized ResNet-32 model from the FACIL codebase has been used. This architecture is not among the standard ResNets introduced in \cite{resnet}. It only features 3 blocks instead of the usual 4 and has been adapted to better handle 32x32 inputs (in terms of feature numbers, kernel sizes, strides, etc.). This architecture has been widely adopted when dealing with CIFAR benchmarks in many previous works in the continual learning area (including \cite{FACIL} and \cite{BIC}) and (as of our knowledge) was first made popular in the iCaRL \cite{icarl} paper. The input image size is 32x32.

For the CORe50-based benchmarks, a variant of the MobileNet v1 \cite{Howard2012} has been used in which the number of intermediate feature maps is limited to 512 (instead of 1024 used in the last 3 layers of the original architecture). The input image size is 128x128 and the model was pretrained on ImageNet.

\bibliography{TPC.bib}
\bibliographystyle{IEEEtran}

\newpage
\section*{Biography}
\vspace{-1cm}
\begin{IEEEbiography}[{\includegraphics[width=1in,height=1.25in,clip,keepaspectratio]{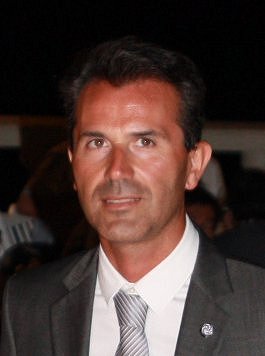}}]{Davide Maltoni}
is a Full Professor at University of Bologna (Dept. of Computer Science and Engineering - DISI). His research interests are in the area of Computer Vision and Machine Learning. Davide Maltoni is co-director of the Biometric Systems Laboratory (BioLab), which is internationally known for its research and publications in the field. Several original techniques have been proposed by BioLab team for fingerprint feature extraction, matching and classification, for hand shape verification, for face location and for performance evaluation of biometric systems. Davide Maltoni is co-author of the Handbook of Fingerprint Recognition published by Springer, 2022 and holds three patents on Fingerprint Recognition. He has been elected IAPR (International Association for Pattern Recognition) Fellow 2010. 
\end{IEEEbiography}

\begin{IEEEbiography}[{\includegraphics[width=1in,height=1.25in,clip,keepaspectratio]{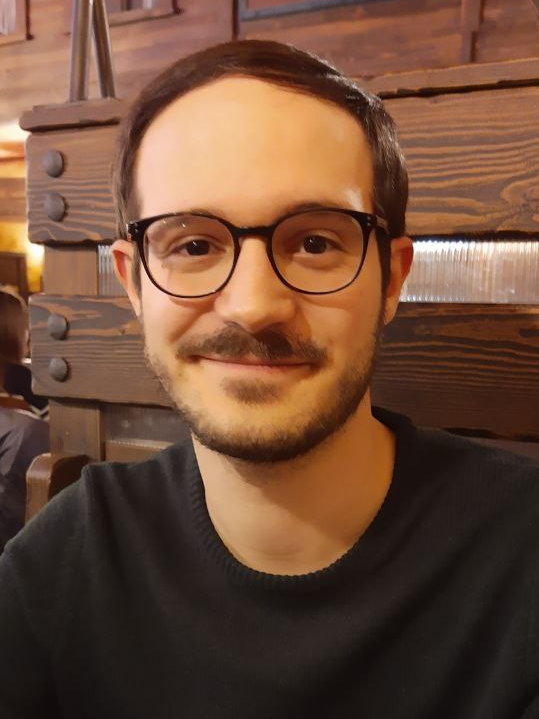}}]{Lorenzo Pellegrini}
is an Assistant Professor at University of Bologna (Dept. of Computer Science and Engineering - DISI). He received his Ph.D. in Computer Science and Engineering at the University of Bologna in 2022 with a dissertation on “Continual Learning for Computer Vision Applications”. He is mainly active in the Computer Vision area, with his main field of interest being Continual Lifelong Learning and Biometric Systems. His main topics include efficient replay mechanisms, bias correction algorithms, and practical Continual Learning applications for mobile and embedded systems. He is a Lead Maintainer for the Avalanche open-source Continual Learning library. He has been a Research Intern at Facebook AI Research in 2021. He has been an organizer of the 3rd and 4th challenges held at the CLVISION workshop at CVPR 2022 and 2023.
\end{IEEEbiography}



\vfill

\end{document}